\documentclass[letterpaper, 10 pt, conference]{ieeeconf}

\IEEEoverridecommandlockouts
\usepackage{amsmath}
\usepackage{amssymb}
\usepackage{booktabs}
\usepackage{graphicx}
\usepackage{multirow}
\usepackage{makecell}
\usepackage{xcolor}

\usepackage{textcomp}
\usepackage{colortbl}  
\usepackage{graphicx}  
\usepackage{float}
\definecolor{uncertainorange}{RGB}{210,105,30}

\usepackage{hyperref}

\title{\LARGE \bf
KFS$^4$LAM: Adaptive KeyFrame Selection with Structure-Aware Mapping for Endoscopic Gaussian Splatting SLAM
}

\author{
Xinwei Ju$^{*}$, Zhehua Mao, Tianyi Song, Danail Stoyanov, Sophia Bano, Francisco Vasconcelos\\
\textit{Hawkes Institute, University College London, London, W1W 7TY, UK}\\
$^{*}$Corresponding author: \texttt{xinwei.ju.22@ucl.ac.uk}
}

\begin{document}

\maketitle
\thispagestyle{empty}
\pagestyle{empty}

\begin{abstract}
Gaussian-splatting Simultaneous Localization and Mapping (SLAM) augments conventional SLAM with a dense, renderable 3D scene representation, enabling high-quality reconstruction and novel-view synthesis. However, its localization, reconstruction, and rendering performance depends strongly on which observations are retained for mapping. Existing systems commonly rely on fixed temporal intervals or simple pose- and overlap-based criteria for keyframe selection. In colonoscopy, highly non-uniform camera motion, repetitive anatomy, specularities, and other visual artifacts can make these criteria unreliable, leading to either redundant keyframes or missed informative viewpoints. We present KFS$^4$LAM (KeyFrame Selection with Structure-aware mapping for endoscopic Gaussian Splatting SLAM), which couples motion-guided adaptive keyframe selection with structurally supervised mapping and restricted temporal deformation. Accumulated motion guides adaptive keyframe selection, with complementary cues suppressing redundant or unreliable candidates. Density-aware mapping with anatomical contour supervision (colon folds) emphasizes structurally informative regions, while restricted temporal deformation accommodates local tissue dynamics without unconstrained geometric drift. We evaluate KFS$^4$LAM against multiple Gaussian-SLAM baselines on C3VD and C3VD-v2, together with cross-backbone and component ablations. KFS$^4$LAM achieves strong overall performance across trajectory, rendering, and depth metrics, with particularly pronounced gains on the more challenging C3VD-v2 dataset. In the matched comparison with SplaTAM on C3VD-v2, it reduces rotational relative pose error by 20.3\% and depth RMSE by 38.4\%, while improving PSNR by 0.78 dB. Cross-backbone experiments further show that the framework is not tied to a single underlying SLAM architecture. Code and visualization results are available at
\href{https://xinweiju.github.io/KFSLAM/}{Project Page}.
\end{abstract}

\section{Introduction}

Accurate camera tracking and dense reconstruction from endoscopic video can support spatial navigation, anatomy/lesion localization, and procedure documentation ~\cite{dunnbeltran2024nflba,muhlethaler2022coverage}.
However, endoscopic scenes differ markedly from the rigid, well-textured environments assumed by conventional SLAM, exhibiting deformable tissue, a restricted field of view, specular reflections, non-uniform illumination, and repetitive or texture-poor anatomy~\cite{dunnbeltran2024nflba}.
Camera motion is also highly non-uniform, causing the information contributed by consecutive frames to vary substantially.
Slow or near-stationary motion produces long runs of near-redundant observations, whereas rapid translation, rotation, or viewpoint change can reveal substantial new scene content within only a few frames.
Consequently, treating observations uniformly can devote mapping effort to redundant views while overlooking more informative ones, making observation selection an important factor in both camera tracking and reconstruction quality.

Compared with earlier implicit neural representations~\cite{nice_slam}, 3D Gaussian Splatting represents the scene using explicit, efficiently renderable anisotropic Gaussians, enabling localized map updates and high-quality rendering, as demonstrated in GaussianSLAM~\cite{yugay2023gaussianslam}, MonoGS~\cite{matsuki2024gaussiansplatting}, and SplaTAM~\cite{keetha2024splatam}. Recent endoscopic methods have improved robustness to deformable tissue and weak texture through endoscopy-specific tracking in EndoGSLAM~\cite{wang2024endogslam}, deformation-aware representations~\cite{zhu2024endogs,liu2024endogaussian}, and stronger geometric regularization such as surface-conforming constraints, normal supervision, and pose-aware keyframe strategies~\cite{huang2025endo2dtam,bonilla2024gaussianpancakes}. However, these advances have primarily focused on tracking, scene representation, and map optimization, with comparatively less attention to adaptive observation selection and the explicit use of persistent anatomical structure during mapping.

\begin{figure*}[t]
    \centering
    \includegraphics[width=\textwidth]{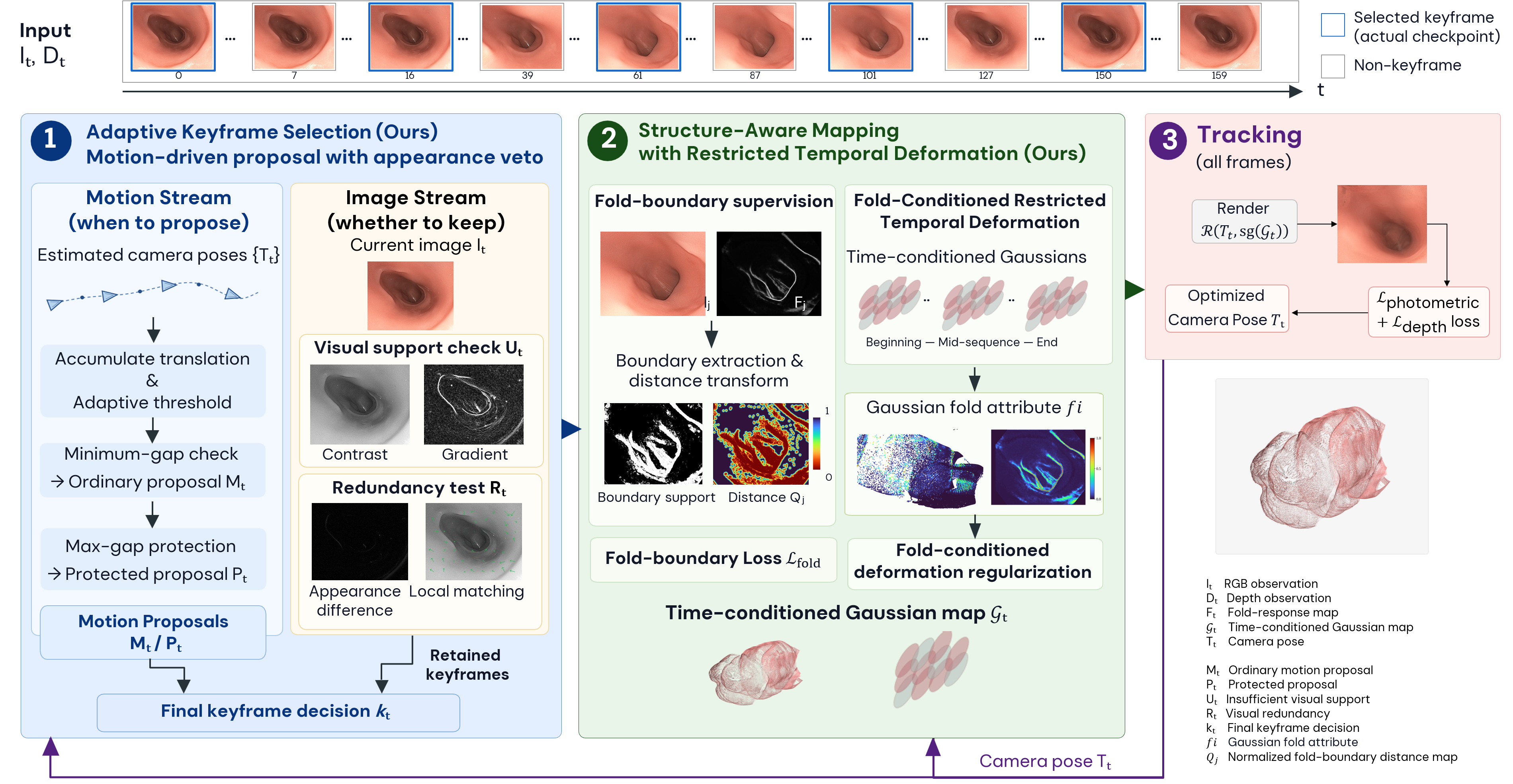}
    \caption{\textbf{Overview of the KFS$^4$LAM pipeline.}
    Given an RGB-D sequence, KFS$^4$LAM consists of three coupled components:
    (1) an adaptive keyframe observation policy that generates candidates from accumulated camera motion and adaptively retains informative, non-redundant observations;
    (2) structure-aware mapping that incorporates fold-boundary supervision and fold-conditioned restricted temporal deformation into the time-conditioned 3D Gaussian map; and
    (3) frame-wise tracking against the learned map.
    Selected keyframes, indicated by blue boxes in the input sequence, provide historical observations for mapping, while all processed frames update the map and are tracked against its current state.
    The resulting representation supports temporally deformable reconstruction while maintaining camera tracking throughout the sequence.}
    \label{fig:pipeline}
\end{figure*}

Keyframe selection in conventional SLAM is commonly based on geometric and visibility cues such as displacement, parallax, and covisibility~\cite{campos2021orbslam3,lsdslam}. Related criteria are also widely used for view selection in structure-from-motion and multi-view reconstruction~\cite{schonberger2016sfmrevisited,schoenberger2016mvs}, while more recent approaches have explored richer geometric and photogrammetric criteria~\cite{pks} and learned, data-driven selection policies~\cite{rl_keyframe_vo}.
In endoscopy, however, assessing viewpoint novelty from visual and correspondence cues remains challenging due to specular reflections, illumination variation, tissue deformation, weak texture, and repetitive anatomy~\cite{ozyoruk2021endoslam,azagra2022endomapper,bobrow2023c3vd}. These effects can produce substantial appearance or correspondence changes without necessarily indicating a geometrically informative new viewpoint~\cite{rota2025endoscopicmatching}.
Existing endoscopic Gaussian-SLAM systems largely rely on temporally spaced keyframes in EndoGSLAM~\cite{wang2024endogslam} or pose/overlap-driven policies in Endo-2DTAM~\cite{huang2025endo2dtam}, with limited consideration of observation redundancy and reliability under non-uniform camera motion. Beyond keyframe selection, anatomical structures such as colon folds can complement appearance-based geometric reasoning~\cite{ju2024segcol,ju2026multimodal} and provide persistent spatial constraints for endoscopic navigation and reconstruction~\cite{dunnbeltran2026centerline}. Incorporating such structure into Gaussian mapping therefore offers a complementary source of supervision to RGB appearance.

Building on these observations, we introduce KFS$^4$LAM, a structure-aware Gaussian-SLAM framework that couples motion-guided adaptive keyframe selection with structurally supervised mapping and restricted temporal deformation. Rather than allowing appearance change to directly trigger keyframe insertion, accumulated camera motion proposes candidate viewpoints, while complementary image- and feature-level cues suppress unreliable or redundant candidates. The resulting non-uniform keyframe density is accommodated by adapting the mapping budget to the number of retained keyframes. The selected observations are further integrated with anatomical fold supervision and restricted temporal deformation, using persistent fold structure to regularize Gaussian mapping while accommodating local non-rigid tissue motion.

Taken together, KFS$^4$LAM provides an adaptive mapping strategy that balances viewpoint informativeness, observation efficiency, and structural consistency in deformable endoscopic scenes. Our evaluation shows consistent improvements in localization and map quality without relying on a higher keyframe density, with the benefits becoming more pronounced on the more challenging C3VD-v2 sequences. Component and cross-backbone ablations further indicate that these gains arise from the proposed design rather than from increased mapping effort or a particular Gaussian-SLAM backbone.

The main contributions of this work are:
\begin{itemize}
    \item A motion-guided adaptive keyframe policy that combines accumulated-motion-based candidate proposal with redundancy-aware visual suppression.

    \item A structure-aware mapping design that combines density-aware mapping with anatomical fold-boundary supervision and fold-conditioned restricted temporal deformation, providing complementary image-space structural supervision and a soft spatial prior for non-rigid Gaussian motion, enabling non-uniform observation allocation under varying camera motion.

    \item A comprehensive evaluation on C3VD and C3VD-v2, including component and matched cross-backbone ablations that isolate the effects of adaptive keyframe selection, structural supervision, and restricted temporal deformation, and demonstrate consistent improvements across different Gaussian-SLAM backbones.
\end{itemize}

\section{Methods}

\subsection{Overview and Problem Formulation}

As illustrated in Fig.~\ref{fig:pipeline}, KFS$^4$LAM couples adaptive keyframe selection with structure-aware mapping for Gaussian-splatting SLAM in deformable endoscopic scenes. The keyframe policy selects observations for subsequent  mapping, while predicted colon folds provide structural supervision and a soft prior for restricted temporal deformation.

Given an RGB-D endoscopic sequence $\{I_t,D_t\}_{t=0}^{T-1}$, spatially aligned fold-response maps $\{F_t\}_{t=0}^{T-1}$, and camera intrinsics $K$, our objective is to estimate camera poses $\{T_t\in\mathrm{SE}(3)\}_{t=0}^{T-1}$ and a canonical Gaussian map $\mathcal G_c=\{g_i\}_{i=1}^{N}$. We retain the RGB-D input setting of the underlying SLAM backbone.
Each $F_t\in[0,1]^{H\times W}$ ($H\times W$ denotes the image resolution) is a continuous per-pixel fold-response map predicted from $I_t$ by a pretrained fold-segmentation model.

Each Gaussian is represented as
\begin{equation}
g_i=(\mu_i,\Sigma_i,c_i,\alpha_i,f_i,B_i),
\end{equation}
where $\mu_i$, $\Sigma_i$, $c_i$, and $\alpha_i$ denote its canonical mean, covariance, colour, and opacity, respectively. The fold attribute $f_i\in[0,1]$ stores the fold response initialized from its source observation when the Gaussian is inserted, while $B_i\in\mathbb R^{3\times2}$ parameterizes restricted temporal displacement. Replacing each canonical mean $\mu_i$ with its time-conditioned mean $\mu_i(t)$ defines the Gaussian map $\mathcal G_t$ used for rendering, mapping, and frame-wise tracking.

\subsection{Adaptive Keyframe Selection}
\label{sec:keyframe_selection}

We separate keyframe selection into motion-based proposal and asymmetric visual suppression: motion generates candidates, while visual cues can only reject uninformative or redundant observations and cannot independently trigger insertion.

\paragraph{Accumulated motion proposal}
Let $\mathbf t_t$ denote the estimated pose translation parameter used by the backbone. We accumulate its frame-to-frame changes as a motion proxy and adapt the proposal threshold to recent motion:
\begin{equation}
\begin{aligned}
d_t &= \|\mathbf t_t-\mathbf t_{t-1}\|_2,\qquad a_t=a_{t-1}+d_t,\\
\theta_t &= \max(d_{\min},g_0\widetilde d_t),
\end{aligned}
\label{eq:motion_proposal}
\end{equation}
where $d_t$ is the frame-to-frame translation change, $a_t$ the accumulated translation since the most recent proposal, $\theta_t$ the adaptive proposal threshold, and $\widetilde d_t$ the median positive translation step over recent 16 frame pairs. We use a reference keyframe interval $g_0=8$ and minimum motion threshold $d_{\min}=0.03$. The accumulator is initialized to zero and reset after every proposal, including those subsequently rejected. 

Let $g_t$ denote the number of processed frames since the last retained keyframe. An ordinary proposal $M_t$ is generated when $a_t\geq\theta_t$ and $g_{\min}\leq g_t<g_{\max}$. A protected proposal $P_t$ is generated when $g_t\geq g_{\max}$, or at the penultimate frame if $g_t\geq g_{\min}$. We use $(g_{\min},g_{\max})=(4,16)$.

\paragraph{Asymmetric visual suppression}
Ordinary proposals are compared with up to three recent retained keyframes. We set $U_t=1$ if the mean Sobel-gradient magnitude or grayscale standard deviation falls below $0.65$ or $0.60$ times the corresponding reference median. We set $R_t=1$ if either of two tests holds: an appearance test with minimum grayscale and gradient-magnitude MAEs of at most $0.025$ and $0.020$, or a pose-free local-matching test with grayscale MAE $\leq0.025$ and normalized mean local-matching displacement $\leq0.10$ for the reference with minimum grayscale MAE, together with maximum gradient cosine similarity $\geq0.92$ over the references.

The final insertion decision is
\begin{equation}
k_t=[t=0]\lor P_t\lor(M_t\land\neg U_t\land\neg R_t).
\label{eq:kfs_gate}
\end{equation}
Protected proposals bypass suppression, while visual cues can only reject ordinary proposals rather than trigger insertion.

\begin{figure*}[!t]
    \centering
    \includegraphics[width=\linewidth]{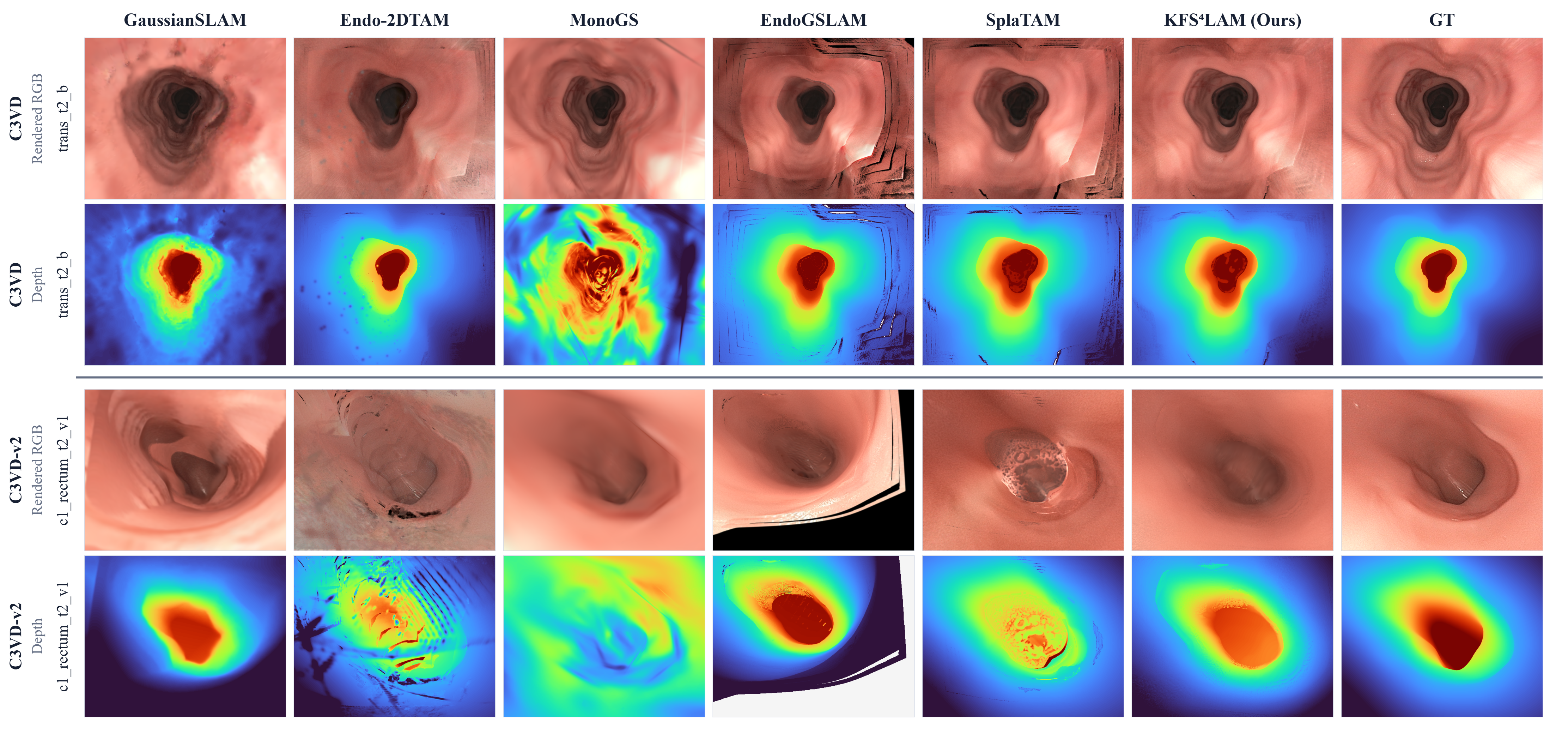}
    \caption{\textbf{Qualitative results on C3VD~\cite{bobrow2023c3vd} and C3VD-v2~\cite{golhar2025c3vdv2}.} We compare our method (KFS$^4$LAM) with the SOTA methods for dense endoscopic SLAM, including MonoGS~\cite{matsuki2024gaussiansplatting}, EndoGSLAM~\cite{wang2024endogslam}, Endo-2DTAM~\cite{huang2025endo2dtam}, GaussianSLAM~\cite{yugay2023gaussianslam}, and SplaTAM~\cite{keetha2024splatam}. Our method generates more robust color and depth reconstruction, as shown by results from trans\_t2\_b (C3VD) and c1\_rectum\_t2\_v1 (C3VD-v2).}
   \label{fig:rendering_results}
\end{figure*}

\subsection{Structure-Aware Restricted Temporal Mapping}

Predicted fold structure provides two complementary signals for mapping: image-space structural supervision and a soft spatial prior for restricted temporal deformation.

\paragraph{Fold-boundary distance supervision}
For a mapping observation indexed by $j$, the stored Gaussian fold attributes $f_i$ are alpha-composited to render the fold-response map $\widehat F_j$. We apply the Sobel boundary-magnitude operator $\mathcal B(\cdot)$ to both the observed fold-response map $F_j$ and the rendered response $\widehat F_j$. We threshold $\mathcal B(F_j)$ to obtain a binary fold-boundary map and compute its Euclidean distance transform. The distance is truncated at $10$ pixels to limit the influence of distant boundaries and normalized by $10$, yielding the normalized distance map $Q_j$. The structural loss and mapping objective are
\begin{equation}
\begin{aligned}
\mathcal L_{\rm fold}
&=\frac{1}{HW}\sum_x \mathcal B(\widehat F_j)(x)\operatorname{sg}(Q_j(x)),\\
\mathcal L_{\rm map}
&=\mathcal L_{\rm map}^{\rm backbone}+\mathcal L_{\rm fold}+\mathcal L_{\rm def}.
\end{aligned}
\label{eq:fold_boundary_dt}
\end{equation}
We refer to $\mathcal L_{\rm fold}$ as the Boundary-DT loss, where $\operatorname{sg}(\cdot)$ denotes stop-gradient and $\mathcal L_{\rm map}^{\rm backbone}$ is the native mapping objective of the underlying backbone. This one-sided loss penalizes rendered boundaries that lie far from observed fold boundaries, without requiring every observed boundary to be matched. Gradients from $\mathcal L_{\rm fold}$ update Gaussian geometry, opacity, and temporal deformation, while the stored fold attributes and mapping pose are detached.

\paragraph{Fold-conditioned temporal deformation}
Each Gaussian receives a low-dimensional time-dependent translation
\begin{equation}
\begin{aligned}
\mu_i(j)
&=\mu_i+\Delta\mu_i(j), \qquad
\Delta\mu_i(j)=B_i
\begin{bmatrix}
u_j\\
\sin(\pi u_j)
\end{bmatrix},\\
u_j
&=\frac{j}{\max(T-1,1)}.
\end{aligned}
\label{eq:temporal_deformation}
\end{equation}
Only the Gaussian mean is time-conditioned. The linear and half-sine terms capture gradual and smooth nonlinear displacement, respectively, while both vanish at the first frame and anchor the canonical geometry. The resulting Gaussians form the time-conditioned map $\mathcal G_j=\{g_i(j)\}_{i=1}^{N}$.

With $N$ denoting the number of Gaussians currently stored in the canonical map, $j^-=\max(j-1,0)$ and $w_i=1-\beta\,\operatorname{sg}(f_i)$, the deformation regularizer is
\begin{equation}
\begin{aligned}
\mathcal L_{\rm def}
&={}\frac{\lambda_{\rm mag}}{N}\sum_i w_i\|\Delta\mu_i(j)\|_2^2 \\
&+\frac{\lambda_{\rm temp}}{N}\sum_i\|\Delta\mu_i(j)-\Delta\mu_i(j^-)\|_2^2\\
&+\lambda_{\rm mean}\left\|\frac{1}{N}\sum_i\Delta\mu_i(j)\right\|_2^2.
\end{aligned}
\label{eq:def_loss}
\end{equation}
We use $\beta=0.5$ and $(\lambda_{\rm mag},\lambda_{\rm temp},\lambda_{\rm mean})=(0.001,0.01,0.01)$. The terms regularize deformation magnitude, temporal variation, and global mean displacement, while higher fold response values relax the magnitude penalty and provide a soft spatial prior.

\paragraph{Tracking with time-conditioned geometry}
At frame $t$, tracking optimizes only the camera pose while keeping the time-conditioned map $\mathcal G_t$ fixed:
\begin{equation}
T_t^\ast=\arg\min_{T_t}\mathcal L_{\rm track}^{\rm backbone}\left(\mathcal R(T_t,\operatorname{sg}(\mathcal G_t)),I_t,D_t\right),
\label{eq:tracking_backbone}
\end{equation}
where $\mathcal R$ denotes Gaussian rendering and $\mathcal L_{\rm track}^{\rm backbone}$ is the native tracking objective. Map and deformation parameters remain fixed during tracking; fold-boundary supervision and deformation regularization are applied only during mapping.

\begin{table*}[!t]
\caption{Quantitative trajectory results of Endo-2DTAM~\cite{huang2025endo2dtam}, EndoGSLAM~\cite{wang2024endogslam}, GaussianSLAM~\cite{yugay2023gaussianslam}, MonoGS~\cite{matsuki2024gaussiansplatting}, SplaTAM~\cite{keetha2024splatam}, and KFS$^4$LAM on C3VD~\cite{bobrow2023c3vd} and C3VD-v2~\cite{golhar2025c3vdv2}. Best and second-best results are shown in bold and underlined, respectively.}
\centering
\renewcommand{\arraystretch}{1.0}
\resizebox{\textwidth}{!}{%
\begin{tabular}{l|rrrrrr|rrrrrr}
\toprule
\multicolumn{1}{c|}{\multirow{2}{*}[-8pt]{Method}}
& \multicolumn{6}{c}{C3VD}
& \multicolumn{6}{c}{C3VD-v2} \\
\cmidrule(lr){2-7}
\cmidrule(lr){8-13}
& \makecell{wATE $\downarrow$\\(mm)}
& \makecell{mATE $\downarrow$\\(mm)}
& \makecell{wARE $\downarrow$\\(\textdegree)}
& \makecell{mARE $\downarrow$\\(\textdegree)}
& \makecell{RPE-t $\downarrow$\\(mm)}
& \makecell{RPE-r $\downarrow$\\(\textdegree)}
& \makecell{wATE $\downarrow$\\(mm)}
& \makecell{mATE $\downarrow$\\(mm)}
& \makecell{wARE $\downarrow$\\(\textdegree)}
& \makecell{mARE $\downarrow$\\(\textdegree)}
& \makecell{RPE-t $\downarrow$\\(mm)}
& \makecell{RPE-r $\downarrow$\\(\textdegree)} \\
\midrule
GaussianSLAM~\cite{yugay2023gaussianslam}
& 3.8531 & 4.0885 & 93.1512 & 77.3899 & 0.3392 & 0.1930
& 10.7119 & 11.2901 & 127.27 & 112.75 & 0.7238 & \textbf{0.4446} \\

MonoGS~\cite{matsuki2024gaussiansplatting}
& 2.4881 & 2.7190 & 57.4049 & 63.4965 & 0.3791 & 0.5640
& 6.2637 & \textbf{7.1637} & 56.3621 & 70.1446 & 0.6002 & 1.4456 \\

Endo-2DTAM~\cite{huang2025endo2dtam}
& 1.4794 & 1.5718 & 42.5380 & 47.2588 & 0.1321 & 0.6578
& 9.8746 & 10.7859 & 85.5424 & 89.1716 & 0.4476 & 2.1940 \\

EndoGSLAM~\cite{wang2024endogslam}
& 0.4773 & 0.5377 & 8.4370 & 10.7657 & 0.0888 & 0.1617
& \underline{6.0724} & 7.5037 & \underline{52.9959} & 60.1639 & 0.4411 & 1.2234 \\

SplaTAM~\cite{keetha2024splatam}
& \underline{0.4637} & \underline{0.4544} & \underline{8.2745} & \underline{10.3404}
& \underline{0.0625} & \underline{0.1464}
& 6.4333 & 7.6836 & 64.1820 & \underline{59.4920} & \textbf{0.3103} & 1.3969 \\

KFS$^4$LAM (Ours)
& \textbf{0.4520} & \textbf{0.4476} & \textbf{7.7085} & \textbf{9.3457}
& \textbf{0.0624} & \textbf{0.1418}
& \textbf{5.8463} & \underline{7.2429} & \textbf{43.1290} & \textbf{44.8062}
& \underline{0.3764} & \underline{1.1134} \\
\bottomrule
\end{tabular}}
\label{tab:trajectory_results}
\end{table*}


\begin{table*}[!t]
\caption{Quantitative rendering results of Endo-2DTAM~\cite{huang2025endo2dtam}, EndoGSLAM~\cite{wang2024endogslam}, GaussianSLAM~\cite{yugay2023gaussianslam}, MonoGS~\cite{matsuki2024gaussiansplatting}, SplaTAM~\cite{keetha2024splatam}, and KFS$^4$LAM on C3VD~\cite{bobrow2023c3vd} and C3VD-v2~\cite{golhar2025c3vdv2}. Best and second-best results are shown in bold and underlined, respectively.}
\centering
\renewcommand{\arraystretch}{1.0}
\begin{tabular}{l| r r r r| r r r r}
\toprule
\multicolumn{1}{c|}{\multirow{2}{*}[-8pt]{Method}} & \multicolumn{4}{c}{C3VD} & \multicolumn{4}{|c}{C3VD-v2} \\
\cmidrule(lr){2-5} \cmidrule(lr){6-9}
& \makecell{PSNR $\uparrow$\\(dB)}
& \makecell{SSIM $\uparrow$}
& \makecell{LPIPS $\downarrow$}
& \makecell{Depth RMSE $\downarrow$\\(mm)}
& \makecell{PSNR $\uparrow$\\(dB)}
& \makecell{SSIM $\uparrow$}
& \makecell{LPIPS $\downarrow$}
& \makecell{Depth RMSE $\downarrow$\\(mm)} \\
\midrule
Endo-2DTAM~\cite{huang2025endo2dtam} & 20.2745 & 0.7867 & 0.3603 & 9.0555 & 18.3235 & 0.6852 & 0.4906 & 13.1850 \\
EndoGSLAM~\cite{wang2024endogslam} & 21.1734 & 0.7319 & 0.3111 & 2.6904 & 14.6158 & 0.5615 & 0.6484 & 14.8925 \\
GaussianSLAM~\cite{yugay2023gaussianslam} & 25.7629 & 0.7498 & 0.5148 & 6.5036 & \underline{22.6829} & 0.6544 & 0.6147 & 10.5528 \\
MonoGS~\cite{matsuki2024gaussiansplatting} & \textbf{29.6784} & 0.8482 & 0.3824 & 1430.08 & \textbf{25.9608} & \textbf{0.7774} & 0.6067 & 1106.53 \\
SplaTAM~\cite{keetha2024splatam} & 25.6437 & \underline{0.8776} & \underline{0.2416} & \underline{1.1509} & 20.1354 & 0.7053 & \textbf{0.4257} & 4.9467 \\
KFS$^4$LAM (Ours) & \underline{26.6164} & \textbf{0.8835} & \textbf{0.2206} & \textbf{1.0720} & 20.9113 & \underline{0.7163} & \underline{0.4332} & \textbf{3.0472} \\
\bottomrule
\end{tabular}
\label{tab:rendering_results}
\end{table*}

\subsection{Optimization and Adaptive Mapping Budget}
To accommodate the non-uniform replay pool induced by adaptive keyframe selection, we adapt the per-frame mapping budget to the number of retained observations. Let $n_t$ denote the number of retained historical keyframes before mapping processed frame $t$. Relative to a fixed-gap reference, we define
\begin{equation}
\begin{aligned}
E_t&=\left\lceil\frac{t+1}{g_0}\right\rceil+1,\qquad
r_t=\max\left(1,\frac{n_t+1}{E_t}\right),\\
N_{\rm map}(t)&=\operatorname{clip}\!\left(
\left\lceil N_0\min(s_{\max},r_t^p)\right\rceil,
N_{\min},N_{\max}\right).
\end{aligned}
\label{eq:mapping_budget}
\end{equation}
Here, $E_t$ is the nominal replay-pool size under fixed-gap insertion, including the current observation, while $n_t+1$ is the available pool size under the adaptive policy. We use $g_0=8$, $N_0=N_{\min}=25$, $N_{\max}=75$, $s_{\max}=3$, and $p=1$ for both backbones. Larger-than-reference replay pools therefore receive additional mapping iterations.

\section{Experimental Setup}

\subsection{Datasets and Baselines}
We evaluate on registered and undistorted RGB-D sequences from C3VD and C3VD-v2, comprising 14 C3VD sequences (5,810 frames) and 10 C3VD-v2 sequences (4,511 frames).

We compare against GaussianSLAM, Endo-2DTAM, MonoGS, EndoGSLAM, and SplaTAM.
KFS$^4$LAM is instantiated on both EndoGSLAM and SplaTAM to evaluate its applicability across reconstruction backbones, while the baselines retain their native tracking, mapping, and keyframe mechanisms.

\subsection{Experimental Settings}
All KFS$^4$LAM variants operate at $675\times540$ resolution.
The EndoGSLAM and SplaTAM baselines use a fixed keyframe interval of 8 processed frames, whereas our adaptive policy uses minimum, reference, and maximum gaps of 4, 8, and 16 frames.
We use 15 tracking iterations and adapt the mapping budget from 25 to 75 iterations.
Fold responses are obtained by reoroducing the best-performing model from~\cite{ju2024segcol}, trained with EndoMapper~\cite{azagra2022endomapper} fold labels, and retained as continuous responses in our pipeline.
Experiments are run on NVIDIA Tesla V100-DGXS GPUs with 32,GB memory.

\subsection{Evaluation Protocol}
We report Sim(3)-aligned ATE and ARE and consecutive-frame RPE-t/RPE-r
in millimeters/degrees.
For ATE and ARE, both macro and sequence-length-weighted averages are reported,
\[
\mathrm{m}=\frac{1}{S}\sum_{s=1}^{S}e_s,\qquad
\mathrm{w}=\frac{\sum_{s=1}^{S}N_s e_s}{\sum_{s=1}^{S}N_s},
\]
where $e_s$ is the sequence-level error and $N_s$ its original frame count.
RPE, PSNR, SSIM, LPIPS-Alex, and depth RMSE are macro-averaged across sequences.
Qualitative evaluation includes RGB, depth, and Sim(3)-aligned spatial reconstruction.

\section{Results}

\subsection{Trajectory Accuracy}
Table~\ref{tab:trajectory_results} compares trajectory accuracy on C3VD and C3VD-v2. On C3VD, KFS$^4$LAM achieves the best performance across all reported trajectory metrics, although the gains over SplaTAM are relatively modest. The advantage becomes more pronounced on C3VD-v2, where KFS$^4$LAM achieves the lowest weighted ATE and both the lowest weighted and macro ARE, while obtaining the second-lowest macro ATE. In particular, weighted ARE decreases from $53.00^\circ$ for EndoGSLAM to $43.13^\circ$. Relative pose metrics show a more mixed trend: SplaTAM retains the lowest RPE-t and GaussianSLAM the lowest RPE-r, with KFS$^4$LAM ranking second on both. Taken together, these results indicate that the main tracking benefit of KFS$^4$LAM lies in improved global trajectory consistency, particularly on the more challenging C3VD-v2 sequences.

\subsection{Rendering and Geometric Reconstruction}
Table~\ref{tab:rendering_results} reports image-space rendering and depth reconstruction quality. On C3VD, KFS$^4$LAM achieves the best SSIM, LPIPS, and depth RMSE, while ranking second in PSNR. On C3VD-v2, its strongest improvement is geometric: compared with SplaTAM, depth RMSE decreases from $4.95$~mm to $3.05$~mm, a reduction of $38.4\%$, while PSNR increases by $0.78$~dB and SSIM also improves. LPIPS remains slightly better for SplaTAM. These results suggest that KFS$^4$LAM primarily improves geometric consistency while maintaining competitive image-space rendering quality.

MonoGS provides a notable contrast: although it achieves the highest PSNR on both datasets, it produces very large depth errors.
This discrepancy indicates that photometric reconstruction quality alone is insufficient to characterize the geometric fidelity of dense endoscopic SLAM, a distinction that is also evident in the qualitative RGB and depth rendering results.

Fig.~\ref{fig:rendering_results} is consistent with the quantitative observations.
On C3VD, several methods produce plausible RGB reconstructions, whereas differences become more apparent in the corresponding depth estimates.
The distinction is stronger on C3VD-v2, where the baselines exhibit localized appearance artifacts, distorted cavity structure, or inconsistent depth, whereas KFS$^4$LAM preserves more coherent cavity geometry across the displayed RGB and depth views relative to the ground truth.
In particular, the comparison with MonoGS illustrates the difference between photometric fidelity and geometric consistency: visually plausible RGB synthesis can coexist with substantially inaccurate depth.

\begin{table*}
\caption{Cross-backbone and component results on C3VD and C3VD-v2.
Green indicates improvement over the corresponding backbone.
KF (\%) is $100\sum_s K_s/\sum_s N_s$, where $K_s$ and $N_s$ are retained
keyframe and input-frame counts; it is not ranked.}
\centering
\renewcommand{\arraystretch}{1.05}
\resizebox{\textwidth}{!}{%
\begin{tabular}{l|rrrrrrrrr|rrrrrrrrr}
\toprule
\multicolumn{1}{c|}{\multirow{2}{*}[-8pt]{Method}}
& \multicolumn{9}{c|}{C3VD}
& \multicolumn{9}{c}{C3VD-v2} \\
\cmidrule(lr){2-10}\cmidrule(lr){11-19}
& m ATE & m ARE & RPE-t & RPE-r & PSNR & SSIM & LPIPS & \shortstack{Depth\\RMSE} & KF (\%)
& m ATE & m ARE & RPE-t & RPE-r & PSNR & SSIM & LPIPS & \shortstack{Depth\\RMSE} & KF (\%) \\
\midrule

EndoGSLAM~\cite{wang2024endogslam}
& \underline{0.538} & \textbf{10.766} & \underline{0.089} & \underline{0.162} & \underline{21.173} & \underline{0.732} & \underline{0.311} & \underline{2.690} & 11.274
& \textbf{7.504} & \underline{60.164} & \underline{0.441} & \underline{1.223} & \underline{14.616} & \underline{0.562} & \underline{0.648} & \underline{14.893} & 11.264 \\

KFS$^4$LAM$_{\mathrm{EndoGSLAM}}$ (Ours)
& \cellcolor{green!30}\textbf{0.528} & \underline{12.196} & \cellcolor{green!30}\textbf{0.087} & \cellcolor{green!30}\textbf{0.161} & \cellcolor{green!30}\textbf{21.762} & \cellcolor{green!30}\textbf{0.738} & \cellcolor{green!30}\textbf{0.301} & \cellcolor{green!30}\textbf{2.588} & 10.964
& \underline{7.625} & \cellcolor{green!30}\textbf{56.009} & \cellcolor{green!30}\textbf{0.428} & \cellcolor{green!30}\textbf{0.927} & \cellcolor{green!30}\textbf{15.301} & \cellcolor{green!30}\textbf{0.598} & \cellcolor{green!30}\textbf{0.608} & \cellcolor{green!30}\textbf{11.928} & 11.059 \\

\cmidrule(lr){1-19}

SplaTAM~\cite{keetha2024splatam}
& 0.454 & 10.340 & 0.063 & 0.146 & 25.644 & 0.878 & 0.242 & 1.151 & 12.857
& 7.684 & 59.492 & \textbf{0.310} & 1.397 & 20.135 & 0.705 & \textbf{0.426} & 4.947 & 12.826 \\

SplaTAM + Adaptive Mapping (DynMap)
& \cellcolor{green!10}0.454 & \cellcolor{green!10}10.318 & 0.065 & 0.153 & \cellcolor{green!10}25.656 & 0.877 & 0.243 & 1.173 & 12.857
& \cellcolor{green!10}7.405 & 64.258 & 0.331 & 1.657 & 19.712 & 0.686 & 0.439 & 5.434 & 12.826 \\

\cmidrule(lr){2-10}\cmidrule(lr){11-19}

SplaTAM + Motion Proposal Only
& \cellcolor{green!10}\textbf{0.439} & \cellcolor{green!10}10.167 & 0.064 & 0.153 & 25.587 & 0.873 & \cellcolor{green!10}\underline{0.240} & 1.168 & 12.392
& \cellcolor{green!10}\textbf{6.370} & 67.134 & 0.338 & \cellcolor{green!10}\underline{1.134} & 19.278 & 0.656 & 0.470 & \cellcolor{green!10}4.582 & 12.826 \\

SplaTAM + Motion-Guided KFS
& 0.456 & \cellcolor{green!10}\textbf{9.266} & 0.064 & 0.153 & 25.465 & 0.873 & 0.245 & \cellcolor{green!10}1.117 & 12.341
& 7.932 & 86.630 & 0.342 & 1.451 & 19.138 & 0.659 & 0.449 & \cellcolor{green!10}4.697 & 12.333 \\

SplaTAM + Structure-Guided Mapping
& \cellcolor{green!10}0.454 & 10.396 & 0.064 & 0.148 & \cellcolor{green!10}25.659 & 0.878 & 0.243 & 1.202 & 12.857
& 7.846 & 86.667 & 0.365 & 1.765 & 18.826 & 0.671 & 0.440 & 5.151 & 12.826 \\

SplaTAM + Structure-Guided + Deformation
& \cellcolor{green!20}\underline{0.439} & \cellcolor{green!20}9.497 & \cellcolor{green!20}\textbf{0.061} & \cellcolor{green!20}\underline{0.145} & \cellcolor{green!20}\textbf{26.852} & \cellcolor{green!20}\textbf{0.891} & \cellcolor{green!20}\textbf{0.221} & \cellcolor{green!20}\textbf{1.071} & 12.857
& \cellcolor{green!20}7.559 & \cellcolor{green!20}\underline{55.199} & \underline{0.326} & \cellcolor{green!20}1.172 & 19.390 & 0.678 & 0.430 & \cellcolor{green!20}\underline{4.061} & 12.826 \\

SplaTAM + Motion-and-Structure-Guided
& \cellcolor{green!20}0.453 & \cellcolor{green!20}10.010 & 0.064 & 0.153 & 25.563 & 0.871 & 0.245 & 1.167 & 12.341
& \cellcolor{green!20}7.503 & 61.233 & 0.409 & \cellcolor{green!20}1.349 & \cellcolor{green!20}\underline{20.776} & \cellcolor{green!20}\textbf{0.721} & \underline{0.429} & \cellcolor{green!20}4.305 & 12.312 \\

\cmidrule(lr){2-10}\cmidrule(lr){11-19}

SplaTAM + MonoGS~\cite{matsuki2024gaussiansplatting} KFS
& \cellcolor{green!10}0.451 & \cellcolor{green!10}9.646 & 0.065 & 0.150 & 25.129 & 0.868 & 0.242 & 1.168 & 5.869
& 7.797 & \cellcolor{green!10}56.867 & 0.360 & 1.438 & 19.651 & 0.676 & 0.448 & 5.292 & 9.209 \\

KFS$^4$LAM$_{\mathrm{SplaTAM}}$ (Ours)
& \cellcolor{green!30}0.448 & \cellcolor{green!30}\underline{9.346} & \cellcolor{green!30}\underline{0.062} & \cellcolor{green!30}\textbf{0.142} & \cellcolor{green!30}\underline{26.616} & \cellcolor{green!30}\underline{0.884} & \cellcolor{green!30}\textbf{0.221} & \cellcolor{green!30}\underline{1.072} & 12.272
& \cellcolor{green!30}\underline{7.243} & \cellcolor{green!30}\textbf{44.806} & 0.376 & \cellcolor{green!30}\textbf{1.113} & \cellcolor{green!30}\textbf{20.911} & \cellcolor{green!30}\underline{0.716} & 0.433 & \cellcolor{green!30}\textbf{3.047} & 12.415 \\
\bottomrule
\end{tabular}%
}
\label{tab:backbone_component}
\end{table*}

\section{Ablation Study}
\subsection{Component Contributions}

The SplaTAM block of Table~\ref{tab:backbone_component} evaluates
\textit{Adaptive Mapping (DynMap)}, \textit{Motion Proposal Only},
\textit{Motion-Guided KFS}, \textit{Structure-Guided Mapping}, its
combination with deformation, and the joint
\textit{Motion-and-Structure-Guided} configuration.
These variants include both adaptive-keyframe and fixed-keyframe controls,
allowing keyframe selection and structure-aware mapping to be examined
separately.

\textit{Adaptive Mapping (DynMap)} remains close to the SplaTAM backbone
on C3VD and does not consistently improve C3VD-v2. Thus, adjusting the
mapping budget alone does not reproduce the gains of the full framework.
\textit{Motion Proposal Only} improves individual trajectory measures,
most notably C3VD-v2 mATE, whereas \textit{Motion-Guided KFS} yields
mixed changes across trajectory and rendering metrics, particularly on
C3VD-v2.

The fixed-keyframe \textit{Structure-Guided Mapping} controls isolate
the mapping components from adaptive selection.
\textit{Structure-Guided Mapping} alone provides limited improvements,
whereas adding deformation substantially improves the
trajectory--reconstruction balance, including mARE, RPE-r, and depth RMSE
on C3VD-v2.
Without deformation, \textit{Motion-and-Structure-Guided} instead improves
C3VD-v2 mATE, PSNR, SSIM, and depth RMSE over
\textit{Structure-Guided Mapping}, while using fewer keyframes.
Combining adaptive selection, structural supervision, and restricted
temporal deformation in the full \textit{KFS$^4$LAM$_{\mathrm{SplaTAM}}$} further reduces C3VD-v2 mATE, mARE, RPE-r, and depth RMSE.
These complementary trends indicate that adaptive keyframe allocation
and restricted deformation contribute differently when coupled with
structural supervision.

Finally, \textit{SplaTAM + MonoGS KFS} uses overlap-based keyframe selection and yields a sparser keyframe set but inconsistent gains, whereas \textit{KFS$^4$LAM$_{\mathrm{SplaTAM}}$} provides more consistent improvements.

\begin{figure}[t]
    \centering
    \includegraphics[width=\linewidth]{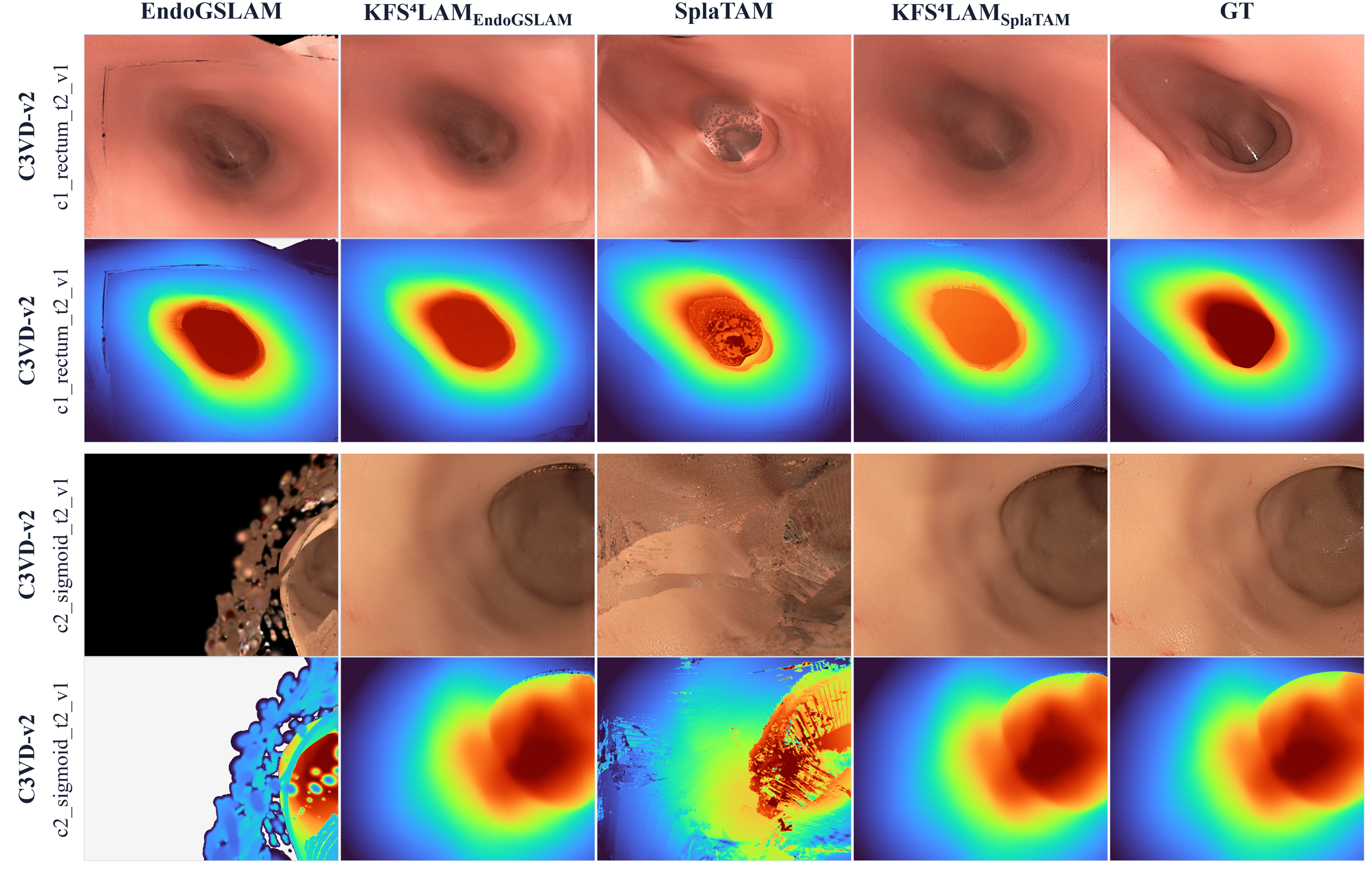}
    \caption{
    \textbf{Rendering generalization across reconstruction backbones on C3VD-v2.}
    Qualitative RGB and depth reconstruction for EndoGSLAM, SplaTAM, and their corresponding KFS$^4$LAM variants on representative C3VD-v2 frames.
    Across both backbones, KFS$^4$LAM produces more coherent cavity structure and suppresses severe appearance and depth artifacts observed in the original reconstructions.
    The improvements are particularly visible on \texttt{c2\_sigmoid\_t2\_v1}, where both KFS$^4$LAM variants recover substantially more consistent RGB and depth structure.
    }
    \label{fig:backbone_rendering}
\end{figure}

\subsection{Keyframe Allocation}

The full-model gains are achieved without a higher keyframe ratio.
KFS$^4$LAM uses a mean keyframe ratio of 12.33\% on C3VD and
12.47\% on C3VD-v2, compared with 13.08\% and 12.90\% for the
fixed-interval SplaTAM baseline.
Table~\ref{tab:backbone_component} also shows lower keyframe ratios for the full EndoGSLAM variant than for its corresponding backbone on both datasets.

Figure~\ref{fig:backbone_trajectory} complements these averages with the temporal distribution of selected observations.
Across both illustrated backbones, the adaptive policy produces locally sparse and dense portions along the sequence, in contrast to the fixed-interval reference policies.
These observations characterize keyframe allocation; the full-model gains are obtained by redistributing a comparable keyframe budget over time rather than by retaining more observations.

\subsection{Cross-Backbone Generalization}

We compare each full KFS$^4$LAM variant with its corresponding SplaTAM~\cite{keetha2024splatam} or EndoGSLAM~\cite{wang2024endogslam} backbone in Table~\ref{tab:backbone_component}.
The keyframe-selection policy, structural supervision, and restricted temporal deformation are transferred without backbone-specific redesign, with only their integration into the native tracking and mapping interfaces differing between the backbones.

On C3VD-v2, the matched SplaTAM comparison gives mARE of $59.49^\circ$ versus $44.81^\circ$ and depth RMSE of $4.95$ versus $3.05$\,mm for the backbone and full model, respectively.
For EndoGSLAM, RPE-r decreases from $1.22^\circ$ to $0.93^\circ$ and depth RMSE from $14.89$ to $11.93$\,mm.
The gains are not uniform across trajectory metrics: SplaTAM has lower RPE-t on C3VD-v2, while EndoGSLAM has lower mATE on C3VD-v2 and lower mARE on C3VD.
The matched comparisons nevertheless support reconstruction benefits across both architectures.

\begin{figure}[h]
\centering
\includegraphics[width=\linewidth]
{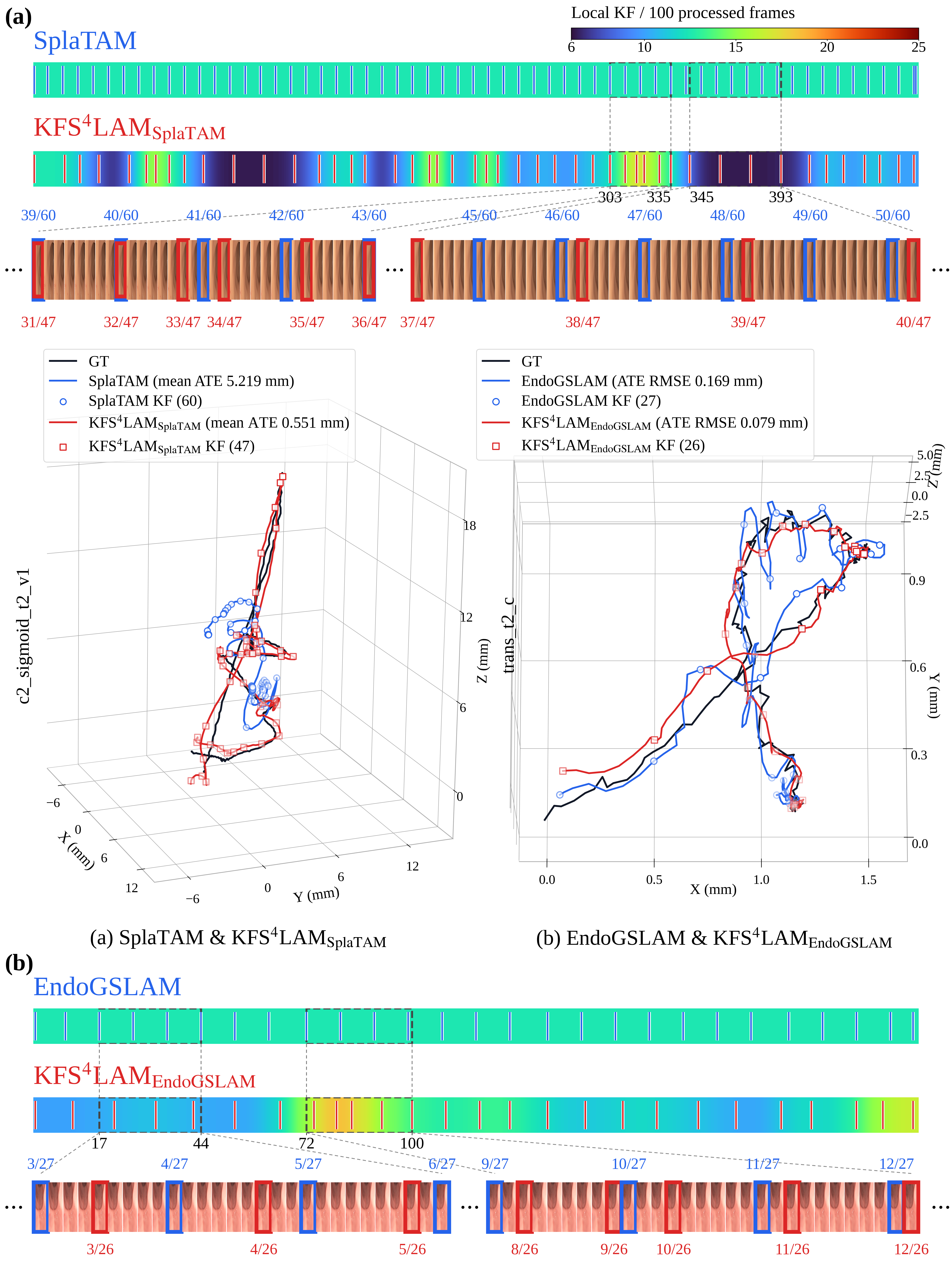}
\caption{\textbf{Cross-backbone generalization of tracking and adaptive keyframe allocation.}
Trajectory and keyframe visualizations of KFS$^4$LAM instantiated with
(a) SplaTAM on C3VD-v2 \texttt{c2\_sigmoid\_t2\_v1} and
(b) EndoGSLAM on C3VD \texttt{trans\_t2\_c}.
Backbone and KFS$^4$LAM trajectories are shown in blue and red,
with circular and square keyframe markers, respectively.
The legends also report ATE for each method and the total number of retained keyframes.
The colored keyframe-density bands visualize smoothed local selection density (keyframes per 100 processed frames) with vertical markers showing individual keyframes; dashed boxes highlight representative sparse and dense periods of 
KFS$^4$LAM and connect them to the corresponding RGB excerpts.
Blue and red boxes identify backbone and KFS$^4$LAM keyframes, respectively, with $x/y$ denoting the $x$-th selected keyframe among the $y$ keyframes retained by that method over the full sequence.}
\label{fig:backbone_trajectory}
\end{figure}

\begin{figure}[t]
    \centering
    \includegraphics[width=\linewidth]{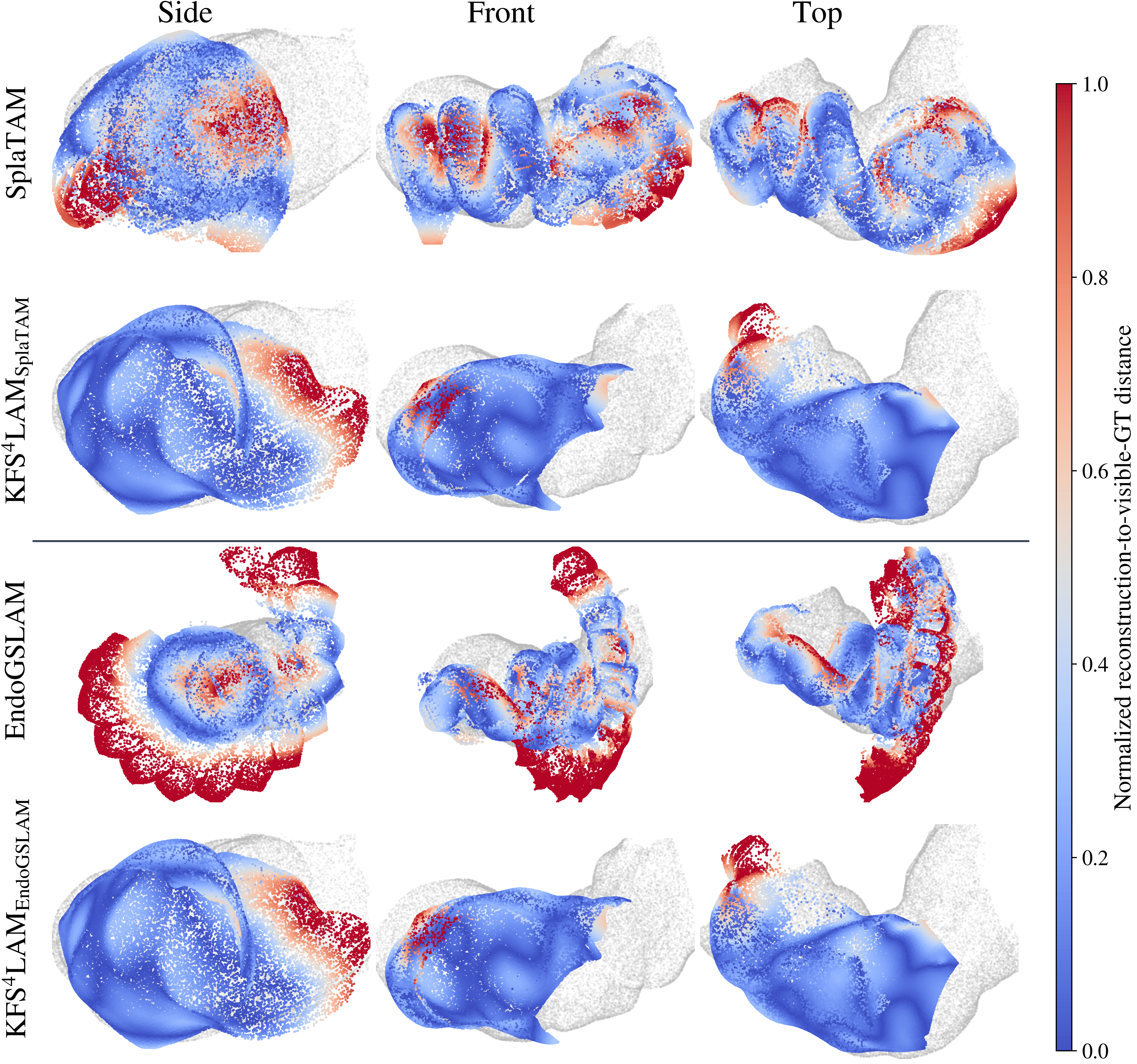}
    \caption{
    \textbf{Geometric generalization across reconstruction backbones on C3VD-v2.}
    Reconstructed Gaussian geometry from EndoGSLAM and SplaTAM, compared with their corresponding KFS$^4$LAM variants.
    Gray denotes the visible ground-truth surface, while color encodes the normalized nearest-neighbor geometric error from the reconstructed Gaussian centers to the visible ground-truth surface.
    Side, front, and top views show that integrating KFS$^4$LAM improves global shape consistency and spatial alignment across both reconstruction backbones.
    }
    \label{fig:backbone_reconstruction}
\end{figure}

Figures~\ref{fig:backbone_rendering} and~\ref{fig:backbone_reconstruction} provide complementary RGB, depth, and spatial evidence, with more coherent cavity structure and improved alignment with the visible reference surface in the illustrated examples.
The corresponding trajectories in Fig.~\ref{fig:backbone_trajectory} further illustrate integration of the observation policy into both tracking pipelines.

\section{Discussion}
\subsection{Interpretation and Evaluation Implications}

The component comparisons suggest that observation selection and map optimization should be designed jointly.
Changing the temporal distribution of retained observations alters the constraints available to mapping, while anatomical supervision and restricted deformation provide complementary mechanisms for maintaining reconstruction consistency.
The controlled ablations further show that structure-aware mapping provides benefits even under fixed keyframe selection, while the full model benefits from combining these components with adaptive observation allocation.

The larger gains on C3VD-v2 may reflect greater sensitivity to observation allocation and scene variation under more challenging acquisition conditions.
Fold-boundary supervision supplies an anatomical constraint beyond appearance and depth, while restricted deformation provides limited flexibility in the recovered geometry.
This is a plausible interpretation of the observed dataset differences, rather than an independently verified explanation of their causes.

The results also motivate joint evaluation of global trajectory consistency, local pose accuracy, image-space rendering, and geometric reconstruction. These measures capture distinct properties of a SLAM system; for example, MonoGS achieves the highest PSNR on both datasets despite substantially larger depth errors. An improvement in one metric therefore cannot be assumed to imply improvement in the others, making the overall trajectory--reconstruction trade-off more informative than a ranking based on a single measure.

\subsection{Limitations}

First, the adaptive mapping budget increases optimization effort when the retained replay pool becomes denser; KFS$^4$LAM should therefore be interpreted as a quality--computation trade-off rather than a runtime improvement.
Although the \textit{DynMap} control isolates budget adaptation, a fixed-compute evaluation would more strictly separate the effects of adaptive mapping and keyframe selection.
Second, fold supervision depends on an external predictor and may become less reliable when anatomical boundaries are weak, occluded, or outside its training distribution.
Finally, although the matched SplaTAM and EndoGSLAM experiments support cross-backbone transfer, evaluation on additional clinical data and more diverse acquisition conditions is required to establish broader generalization.

\section{Conclusion}

We presented KFS$^4$LAM, which couples adaptive observation selection with structure-aware mapping for deformable endoscopic Gaussian SLAM. Experiments on C3VD and C3VD-v2 show that reallocating keyframes according to motion and observation reliability, together with anatomical supervision and restricted deformation, improves the trajectory--reconstruction trade-off across different Gaussian-SLAM backbones without increasing keyframe density. Matched experiments with SplaTAM and EndoGSLAM further show that these benefits are not specific to a single Gaussian-SLAM backbone.


More broadly, our results suggest that keyframe selection in endoscopic SLAM should not be treated independently from the mapping process: changing which observations are retained also changes the spatial and temporal constraints available for reconstruction. The pronounced improvements on C3VD-v2 further indicate that adaptive observation allocation may become increasingly important as motion and scene variation become less uniform. At the same time, the dependence on externally predicted anatomical structure, adaptive mapping computation, and a restricted deformation model limits the current formulation. Future work should investigate jointly optimized selection and mapping under explicit computational budgets, more expressive yet anatomically regularized deformation models, and validation across broader clinical datasets and acquisition conditions.

\bibliographystyle{IEEEtran}
\bibliography{references}

\end{document}